\documentclass[letterpaper]{article}

\usepackage{PRIMEarxiv}

\usepackage[utf8]{inputenc} % allow utf-8 input
\usepackage[T1]{fontenc}    % use 8-bit T1 fonts
\usepackage{booktabs}       % professional-quality tables
\usepackage{amsfonts}       % blackboard math symbols
\usepackage{nicefrac}       % compact symbols for 1/2, etc.
\usepackage{microtype}      % microtypography
\usepackage{lipsum}
\usepackage{fancyhdr}       % header
\usepackage{graphicx}       % graphics
\usepackage{times}  % DO NOT CHANGE THIS
\usepackage{helvet}  % DO NOT CHANGE THIS
\usepackage{courier}  % DO NOT CHANGE THIS
\usepackage[hyphens]{url}  % DO NOT CHANGE THIS
\usepackage{graphicx} % DO NOT CHANGE THIS
\usepackage[round]{natbib}  % DO NOT CHANGE THIS AND DO NOT ADD ANY OPTIONS TO IT
\usepackage{caption} % DO NOT CHANGE THIS AND DO NOT ADD ANY OPTIONS TO IT
\usepackage{algorithm}
\usepackage{algorithmic}

\usepackage{newfloat}
\usepackage{listings}
\DeclareCaptionStyle{ruled}{labelfont=normalfont,labelsep=colon,strut=off} % DO NOT CHANGE THIS
\floatstyle{ruled}
\newfloat{listing}{tb}{lst}{}
\floatname{listing}{Listing}
\usepackage[utf8]{inputenc}
\usepackage{amsmath} % added KB
\usepackage{comment} % added KB
\usepackage{amssymb} % added KB
\usepackage{xcolor}
\usepackage[page]{appendix}

\DeclareMathOperator{\indep}{\perp\!\!\!\perp}
\DeclareMathOperator{\dep}{\not\! \perp\!\!\!\perp}

\title{Conditional Feature Importance with Generative Modeling Using Adversarial Random Forests}
\author{
    Kristin Blesch\textsuperscript{*,\rm 1,\rm2},
    Niklas Koenen\textsuperscript{*,\rm 1,\rm2},
    Jan Kapar\textsuperscript{\rm 1,\rm2},
    Pegah Golchian\textsuperscript{\rm 1,\rm2},\\
   \textbf{Lukas Burk\textsuperscript{\rm 1,\rm2},
    Markus Loecher\textsuperscript{\textdagger, \rm 3},
    Marvin N. Wright}\textsuperscript{\textdagger, \rm 1,\rm2,\rm4} \\[0.75em]
    \textsuperscript{\rm 1}Leibniz Institute for Prevention Research \& Epidemiology – BIPS, Germany  \\
    \textsuperscript{\rm 2}Faculty of Mathematics and Computer Science, University of Bremen, Germany\\
    \textsuperscript{\rm 3}Department of Business and Economics, Berlin School of Economics and Law, Germany\\
    \textsuperscript{\rm 4}Department of Public Health, University of Copenhagen, Denmark\\[0.5em]
    \{blesch, koenen, kapar, golchian, burk, wright\}@leibniz-bips.de, markus.loecher@hwr-berlin.de
}

\begin{document}
\maketitle
\begin{abstract}
This paper proposes a method for measuring conditional feature importance via generative modeling. In explainable artificial intelligence (XAI), conditional feature importance assesses the impact of a feature on a prediction model's performance given the information of other features. Model-agnostic post hoc methods to do so typically evaluate changes in the predictive performance under on-manifold feature value manipulations. Such procedures require creating feature values that respect conditional feature distributions, which can be challenging in practice. Recent advancements in generative modeling can facilitate this. For tabular data, which may consist of both categorical and continuous features, the adversarial random forest (ARF) stands out as a generative model that can generate on-manifold data points without requiring intensive tuning efforts or computational resources, making it a promising candidate model for subroutines in XAI methods. This paper proposes cARFi (conditional ARF feature importance), a method for measuring conditional feature importance through feature values sampled from ARF-estimated conditional distributions. cARFi requires only little tuning to yield robust importance scores that can flexibly adapt for conditional or marginal notions of feature importance, including straightforward extensions to condition on feature subsets and allows for inferring the significance of feature importances through statistical tests.
\end{abstract}

\textbf{Code --- }{https://github.com/bips-hb/cARFi\_paper}
%\textbf{OpenReview --- }{https://openreview.net/forum?id=Q5kFE2xIEp}
\renewcommand{\thefootnote}{\fnsymbol{footnote}} 
\footnotetext[1]{These authors contributed equally.}
\footnotetext[2]{M. Loecher and M.N.Wright share last authorship.}
\renewcommand{\thefootnote}{\arabic{footnote}}

%\keywords{XAI \and Conditional Feature Importance \and  Generative Modeling \and Adversarial Random Forest}

\section{Introduction} \label{sec::intro}
% What is feature importance measurement
Explainable artificial intelligence (XAI) aims to shed light on the opaque behavior of machine learning algorithms, which includes assessing the importance of features for a predictive algorithm. Model-agnostic post hoc methods attribute scores to input features according to their relevance for the prediction in an arbitrary, already fitted supervised machine learning model \citep{molnar2020, murdoch2019}. Refined conceptualizations include, for example, methods aiming for insights on the prediction of individual observations, like Shapley additive explanations \citep{lundberg2017}, or a feature importance focus on the model’s overall behavior, yielding global-level explanations. 

% conditional feature importance measurement 
A crucial distinction in feature importance concepts is between conditional and marginal viewpoints \citep{strobl2008, watson2021}: Marginal feature importance evaluates a feature’s impact irrespective of other features included in the model, whereas conditional feature importance takes the predictive information of other features into account. The presence of dependency structures, which real-world datasets frequently exhibit, plays a pivotal role in this distinction because a feature’s impact on the prediction \textit{given}, i.e., on top of the predictive information provided by correlated features, alters the importance score attributed \citep{watson2021}. To that end, it is worth highlighting that also in-between measures exist, e.g., conditioning on only few features as in relative feature importance \citep{koenig2021} or having a parameter that alters the "strictness" of the conditional distribution regarding the degree of extrapolation allowed (see further our discussion on tree-depth within cARFi in Secs.~\ref{sec::methods} and \ref{sec::proof-of-concept}). Overlooking such nuances poses a pitfall that can result in misinterpretations \citep{molnar2022}. However, determining marginal, conditional, or in-between notions of feature importance in practice necessitates suitable and ready-to-use methods.

% focus of this paper & how do model-agnostic post hoc methods work typically
Especially methods for model-agnostic post hoc conditional feature importance measurement on a global level of explanation urge for improvement. Recap that such methods aim to evaluate the change in a model’s performance when erasing the feature of interest’s predictive information from the dataset, while accounting for the information provided by other features \citep{fisher2019, watson2021}. Approaches may suggest removing the feature of interest from the model altogether, but this commonly involves computationally expensive model refits \citep{lei2018}. More commonly, approaches rely on strategies that remove the dependency on the target while maintaining the features' joint distributional behavior. These methods aim to evaluate the change in predictive performance when replacing the feature of interest's values $x_j$ with $\Tilde{x}_j$ that respect the conditional distribution $p(x_j | \mathbf{X}_{C} = \mathbf{x}_C)$, where $\mathbf{X}_{C}$ indicates the features to condition on (typically all features in the model except for $X_j$). To that end, conditional feature importance approaches may assume that the data is separable into subgroups \citep{molnar2023} or draw on related frameworks such as knockoffs \citep{watson2021} to circumvent the direct modeling of  $p(x_j | \mathbf{X}_{C} = \mathbf{x}_C)$, which is challenging in practice. 

% How generative modeling can help
Generative modeling involves generating new data samples by learning the joint distribution of $\mathbf{X}$. Beyond that, some generative models have the ability to also derive (arbitrary) conditional densities $p(x_j | \mathbf{X}_{C} = \mathbf{x}_C)$ from the joint density without model retraining, which is particularly useful when customizing conditioning sets, e.g., as in relative feature importance. Further, generative models enable the sampling of large amounts of data, which can increase the robustness of feature importance scores. However, generative models are typically computationally intensive and require high effort in parameter tuning \citep{goodfellow2014, xu2019}. To leverage the potential of generative models as a subroutine in feature importance measurement for a broader application range, it is crucial to focus on generative models suitable for continuous and categorical features (mixed data), which require minimal tuning and computational resources. 

%the method
This paper combines recent advancements in generative modeling with XAI and introduces conditional ARF feature importance (cARFi), a method for conditional feature importance measurement using adversarial random forests \citep[ARF,][]{watson2023}. An ARF is an off-the-shelf generative model that is particularly suitable for mixed tabular data, executing in due runtime with comparably little tuning efforts, and allows to efficiently estimate and sample from conditional densities. cARFi is robust, flexibly adjustable for various conditioning sets and even allows for applying statistically valid inference to test for nonzero feature importance.  

%structure
The remainder of this paper is structured as follows: Sec.~\ref{sec::background} introduces relevant background and related work. Then, in Sec.~\ref{sec::methods}, the proposed method is described in more depth and theoretical properties are discussed. Sec.~\ref{sec::evaluation} contains a proof of concept, followed by a simulation study and real data example to evaluate the performance of cARFi under different conditioning sets. Finally, we conclude our findings (Sec.~\ref{sec::conclusion}) and discuss results in Sec.~\ref{sec::discussion}.

\section{Background and Related Work} \label{sec::background}
\subsubsection{Conditional Feature Importance}
% something about conditional feature imoportance
The measurement of conditional feature importance can be approached from various standpoints. Frameworks may focus on leveraging model-specific traits \cite{strobl2008}, study the topic across entire model classes \citep{fisher2019}, or compare model refits that omit the feature of interest \citep[leave-one-covatiate out; LOCO,][]{lei2018}. This paper focuses on model-agnostic methods for a single given model, for which several approaches exist.

% related relevant methods
SAGE values \citep{covert2020} draw on the game-theoretic concept of Shapley values and, in theory, can measure global conditional feature importance by analyzing a feature’s additional contribution across various feature subgroups. In practice, however, approximations that rely on marginal sampling, as in KernelSHAP \cite{lundberg2017}, are frequently used instead of the challenging sampling of values from conditional distributions. As a consequence, this turns the method into a marginal feature importance measure. 

Another approach is analyzing the effect of perturbing feature values within conditional subgroups \citep[CS,][]{molnar2023}. CS works analogous to permutation feature importance \citep[PFI,][]{breiman2001}, but respects the data manifold. The method requires the data to be separable in suitable subgroups that inherent sufficient amounts of data, for which \citet{molnar2023} suggest a tree-based search procedure. While CS, in principle, can also work with conditioning on feature subsets, it does require rerunning the entire procedure for each conditioning set of features. Further, CS evaluates permutations of feature values within subgroups but does not allow for estimating and sampling from the respective conditional distributions. 

A method that, in contrast to SAGE and CS, accommodates a procedure for testing nonzero feature importance is the conditional predictive impact \citep[CPI,][]{watson2021}. This method relies on model-X knockoffs \citep{candes2018}, which are synthetic data samples that imitate the original data and satisfy desirable properties. While knockoff samplers are also generative models, they impose additional properties on the synthesized data, such as equality in the joint distribution of original and generated data under swapping operations \citep{candes2018}, which may be overly strict, and hence disadvantageous for certain tasks \citep{blesch2023unfooling}. Further, knockoff-based procedures may yield unstable results and multiple knockoff imputation might be favorable \citep{gimenez2019}, yet computationally intensive (see further Sec.~\ref{sec::methods}).

% issue: robustness of the procedures
Apparent stumbling blocks of the conditional feature importance methods discussed above are (1) access to conditional distributions and (2) ensuring that sufficient amounts of data are available either in the subgroups or generated by knockoffs, to calculate meaningful and robust feature attributions. cARFi tackles both aspects simultaneously, allowing for the generation of multiple values from conditional distributions in due time. Further, cARFi allows for selecting a hyperparameter that balances the strictness of the conditioning set, making it a flexible measure for both conditional, marginal, and in-between feature importance. 

\subsubsection{Relative Feature Importance}
% issue: focus on relative feature importance often neglected, but very important
In relative feature importance, the evaluation is conditional on \textit{few} other features in the model \citep{koenig2021}. In a certain sense, this concept tones down conditional feature importance which accounts for \textit{all} other features in the model.

Even though relative feature importance is barely discussed in XAI, this concept is highly relevant in practice. For example, if the underlying causal structure includes a collider, e.g., both $X$ and $Y$ share a common effect on $Z$ in the causal graph $X \rightarrow Z \leftarrow Y$. Conditioning on the collider $Z$ creates a non-causal association between $X$ and $Y$, which can mislead interpretations. Similarly, features may be considered "good" or "bad" controls, which should or should not be included as control variables in a model \citep{cinelli2022crash}. Translating this to XAI, users may want to control for only few features, for example, if they are known to be confounding features, but ignore the effects of other features on the importance. 

Conditional feature importance measures can adapt to the concept of relative importance, yet current methods are impractical for this. For example, CS needs to find updated subgroups, and knockoff samplers within CPI must be refit for altered conditioning sets due to the knockoff’s properties. As noted in \citet{koenig2021}, any procedure yielding feature values from the respective conditional distribution works as a subroutine in relative feature importance measurement and may even respect features absent during model training. The challenge, however, is finding a synthesizer that does not rely on strict parametric assumptions, requires refits for changing conditioning sets, or involves vast computational and tuning resources for generating data generally. With recent advancements in generative modeling, this has become feasible and the method proposed in this paper can adapt flexibly to customized conditioning sets without requiring auxiliary calculations.

\subsubsection{Generative Modeling} 
The field of generative modeling is concerned with building models that can generate data instances $\Tilde{\mathbf{X}}$ that follow the same joint distribution as some given data matrix $\mathbf{X}$. Generative models rapidly advance a variety of machine learning-related tasks, such as text generation with large language models \citep{openai2023}, and offer promising lines of research for XAI. 

Tabular data has unique characteristics, such as mixed features, that require careful consideration \citep{borisov2022}. Typically, generative models are based on deep learning architectures, such as in generative adversarial networks \citep{goodfellow2014}, normalizing flows \citep{rezende2015}, variational autoencoders \citep{kingma2013} and diffusion probabilistic models \citep{ho2020}, often combined with transformer-based architectures \citep{vaswani2017}; for an overview, see \citet{bond2021, foster2022}. Such architectures can be difficult for reaching convergence, and are typically computationally demanding and tuning intensive. Adaptions to tabular data exist \citep{xu2019},  yet, tree-based methods may be better suited for tabular data since they require little tuning and naturally handle mixed features \citep{borisov2022, grinsztajn2022}. Attempts to generate data with trees in a more convenient, straightforward manner are thus promising and several such methods have been proposed recently \citep{correia2020, nock2023}.

From such approaches, ARF \citep{watson2023} stands out as a fast and off-the-shelf method for generating high-quality tabular data, requiring only few efforts in tuning and computational resources. Further, it allows for estimating and sampling from the joint as well as conditional distributions, which is essential for the task at hand. The unconditional ARF procedure works as follows:
\begin{enumerate}
    \item Fit unsupervised random forest \citep{shi2006}: First, permute feature values in the given dataset $\mathbf{X}$ randomly across instances to create naive synthetic dataset $\Tilde{\mathbf{X}}$. Then, fit a random forest $\hat{f}^0$ to distinguish instances from $\mathbf{X}$ and $\Tilde{\mathbf{X}}$ (labeled accordingly), where splits in the forest's trees pick up the data's dependency structure.
    \item  If the accuracy of $\hat{f}^0$ is above $50\%$, new synthetic data is sampled from the leaves of forest $\hat{f}^0$ (generator step) and a new random forest $\hat{f}^1$ is fit to classify real and synthetic data (discriminator step).
    \item Data generation and discrimination is continued for $k$ iterations until the accuracy of $\hat{f}^k$ drops down to $50\%$ or below. This indicates that the algorithm has converged, implying that all feature dependencies have been learned and features are mutually independent in the leaves. 
    \item FORDE step (density estimation): The estimated joint density $\hat{p}_\text{ARF}$ can -- thanks to the mutual independence assumption of features within the leaves -- be formulated as a mixture of products $\hat{p}_l$ of univariate densities $\hat{p}_{lj}$ for leaf $l$ and feature $j$, which can be estimated with any arbitrary univariate density estimator within the random forest's leaves, weighted by the share of real data $\pi_l$ that falls into $l$:
    \begin{equation}\label{eq:arf_dens}
        \hat{p}_{\text{ARF}}(\mathbf{x}) = \sum_{l} \pi_l\, \hat{p}_l (\mathbf{x}) = \sum_{l} \pi_l \prod_{j} \hat{p}_{lj}(x_j).
    \end{equation}
    \item FORGE step (data generation): Synthetic data is generated by drawing a leaf $l$ from the random forest with probability $\pi_l$ and then sampling from the estimated univariate densities $\hat{p}_{lj}$ within that leaf.
\end{enumerate}
An ARF model fitted once can generate arbitrary amounts of data from the estimated joint distribution $\hat{p}_{\text{ARF}}(\mathbf{x})$ and derived conditional distributions, as further discussed in Sec.~\ref{sec::methods}. This enables generating multiple values from arbitrary conditional distributions feasible in due time,  which positions ARF as particularly suitable to serve as a subroutine in conditional and relative feature importance measurement. 

\section{Methods} \label{sec::methods}
We propose conditional ARF feature importance (cARFi), a robust measure for conditional and relative feature importance that leverages recent advancements in generative modeling to XAI. cARFi relies on the concept of feature importance measurement that links it to changes in the performance of the prediction model $\hat{f}$ \citep{breiman2001, covert2020,watson2021}. To attribute importance score FI$_j$ to the feature of interest $X_j$, we evaluate the difference in expected loss $\ell$ when removing the effect of $X_j$ on the target $Y$ using the modified dataset $\mathbf{X}^*$. That is, FI$_j$ evaluates $\mathbb{E} [ \ell(\hat{f}(\mathbf{X}^*), Y) ] - \mathbb{E} [ \ell(\hat{f}(\mathbf{X}), Y) ]$. This quantity can be approximated empirically by averaging the instance-wise loss differences across the entire dataset $\{(\mathbf{x}^{(i)}, y^{(i)})\}_{i=1}^N$:
\begin{equation}
\label{eq:pfi}
\widehat{\text{FI}}_j = \frac{1}{N} \sum_{i=1}^{N} \ell\left(\hat{f}(\mathbf{x}^{*(i)}), y^{(i)}\right)  - \ell\left(\hat{f}(\mathbf{x}^{(i)}), y^{(i)}\right).
\end{equation}
Recap that $\hat{f}$ typically cannot handle missing values directly, hence, we have to ensure that the dimension of $\mathbf{X}^*$ matches that of $\mathbf{X}$ instead of analyzing $\mathbf{X}^* := \mathbf{X} \!\setminus\! X_j$ directly. To remove the information contained in $X_j$ on the target $Y$, we therefore aim to replace values of $X_j$ with $\Tilde{x}_j$ such that $\Tilde{X}_j$ is independent of $Y$. While marginal feature importance focuses on $\Tilde{X}_j \indep \{Y, \mathbf{X} \!\setminus\! X_j \}$, for conditional and relative feature importance, this independence must be present considering the features in a conditioning set $\mathbf{X}_C$, i.e., $\Tilde{X}_j \indep Y | \mathbf{X}_C$. Further, the conditional distributions of $X_j | \mathbf{X}_C$ and $\Tilde{X}_j | \mathbf{X}_C$ need to be equal such that the generated data instances are on the data manifold. In sum, we aim for sampling values $\Tilde{x}_j^{(i)}  \sim p(x_j | \mathbf{X}_C = \mathbf{x}_C) $ that satisfy $\Tilde{X}_j \indep Y | \mathbf{X}_C$.

\subsubsection{Conditional ARF Feature Importance}
\label{seq:methods_cARFi}

An ARF can generate such values straightforwardly because of its ability to derive conditional distributions $p(x_j | \mathbf{X}_C = \mathbf{x}_C)$ from the learned joint distribution, with conditional independence to $Y$ satisfied automatically as long as the ARF is fit with only $\mathbf{X}$ (excluding $Y$). As discussed similarly in \citet{dandl2024}, we can leverage the unique traits of ARFs for conditional and relative feature importance measurement as follows: 
\begin{itemize}
    \item Once $\hat{p}_\text{ARF}$ is estimated, ARF allows us to derive estimated conditional densities $\hat{p}_\text{ARF}(x_j | \mathbf{X}_C = \mathbf{x}_C)$ for fixed values $\mathbf{x}_C$ with arbitrary conditioning sets $C$ without the need of refitting the ARF: 
    \begin{equation}
        \hat{p}_\text{ARF}(x_j | \mathbf{X}_C = \mathbf{x}_C) = \sum_{l} \pi_l'\, \hat{p}_{lj}(x_j)
    \end{equation}
    with updated weights $\pi_l' := \pi_l\frac{\hat{p}_l(\mathbf{x}_C)}{\hat{p}_\text{ARF}(\mathbf{x}_C)}$. This includes the case of $C = \{1, \dots, p\} \!\setminus\! \{j\}$ for any feature of interest $j$ and for arbitrary subsets of features $\mathbf{X}_S$, $S \subset \{1, \dots, p\}$ to condition on, as required in relative feature importance measurement. 
    \item With the conditional densities derived, arbitrary amounts of feature values can be sampled in due time as this does not require rerunning the ARF procedure. We can use this computationally cheap sampling to increase the stability (robustness) of the feature importance measure: we propose to sample $R$ values of $\Tilde{x}_j^{(i)}$ for each observation $i$, averaging the change in loss across those replicates before calculating the model-wide feature importance score across all $N$ observations. 
\end{itemize}
In summary, the cARFi method extends the PFI framework by utilizing repeated conditional sampling with ARF. The whole procedure is described in Algorithm~\ref{alg::cpi_seq}.

\begin{algorithm}
\caption{cARFi}\label{alg::cpi_seq}
\textbf{Input}: $(\mathbf{X}^\text{train}, Y^\text{train}), (\mathbf{X}^\text{test}, Y^\text{test})$, learner $f$, feature (set) of interest $j$, conditioning set $C$, ARF procedure $a$, loss function $\ell$, number of replicates $R$
\begin{algorithmic}[1]
%\REQUIRE{$(X^{train}, Y^{train}), (X^{test}, Y^{test})$, supervised learner $f$, feature (or group) of interest $j$, ARF procedure $a$, loss function $L$, number of imputation replicates $R$ }
\STATE learn $\hat{f} \gets f(\mathbf{X}^\text{train}, Y^\text{train})$
\STATE fit ARF $\hat{a} \gets a(\mathbf{X}^\text{train})$ and estimate density $\hat{p}_{\hat{a}}$
\STATE sample $R$  feature values for each test instance $i$:

for each $i \in [N], r \in [R]$: $\Tilde{\mathbf{X}}_j^{\text{test},i(r)} \sim \hat{p}_{\hat{a}}(x_j|\mathbf{X}_{C}^{\text{test},i})$
\STATE define $\Tilde{\mathbf{X}}^{\text{test},i(r)} := \{\Tilde{\mathbf{X}}^{\text{test},i (r)}_{j}, \mathbf{X}^{\text{test},i}_{-j}\}$ and calculate instance-wise loss difference w.r.t. $j$:

$\Delta^{i}_{j} \gets \frac{1}{R} \sum\limits_{r=1}^{R} \ell(\hat{f}(\tilde{\mathbf{X}}^{\text{test},i(r)}, Y))- \ell(\hat{f}(\mathbf{X}^{\text{test},i}, Y))$
\STATE calculate $\widehat{\text{cARFi}}_j \gets \frac{1}{N} \sum_{i=1}^N \Delta^{i}_{j}$

\ENSURE{$\widehat{\text{cARFi}}_j$}
\end{algorithmic}
\end{algorithm}

\subsubsection{Effect of Hyperparameters}
\label{sec:methods_hyperhyper}

Since ARFs solely approximate dependencies across features by their random forests' splits, as reflected in the local independence assumption in Equation~\eqref{eq:arf_dens}, the tree depth has decisive implications. Growing deep trees is desirable to learn the correlations present in the real data accurately and thus yield apt estimates for the underlying joint and conditional data distributions. However, the resulting hyperrectangles defined by the random forest's leaves matching the conditions might become very small, leading to only little variation in feature values sampled when calculating cARFi. Consequently, the changes in the loss might become minimal, resulting in lowered power of statistical tests indicating important features. On the contrary, when growing shallow trees, which yield large hyperrectangles, the estimated conditional distributions might not approximate the real distributions well and the statistical test applied might reject too many null hypotheses, inflating the type I error possibly above the predefined significance level. 

The main parameter controlling for this phenomenon is the minimum node size, i.e., the minimum amount of data points that have to be contained in the terminal nodes of the ARF's trees. In a simulation study, we show how altering this parameter shifts the distribution under the null hypothesis (see Sec.~\ref{sec::proof-of-concept} and Supplement~S2.1). This provides a new perspective on feature importance measurement since it offers the opportunity of in-between feature importance. That is, the notion of marginal and conditional feature importance can be shifted effortlessly by adjusting this parameter, offering new pathways for feature importance measurement.

Another important parameter for this is the finite bounds argument, that, when set to 'local', replaces infinite bound values with the empirical bounds within the leaves. This takes into account the poor extrapolation ability of random forests and helps to keep the generated data on the manifold.

\subsubsection{Statistical Testing}
\label{sec:methods_test}

A crucial property of the cARFi estimator is that it is asymptotically normally distributed, thus providing valid statistical inference for the importance of a set of features $S$ conditioned on a set of other features $C$. The derivation is based on the detailed explanations for the CPI concept proposed by \citet{watson2021} and utilizes the fact that the empirical risk estimator is asymptotically normally distributed. For a loss function $\ell$ that acts instance-wise, we define the following random variable:
\begin{align}
\Delta = \frac{1}{R} \sum_{r = 1}^{R} \ell\left(\hat{f}(\{\tilde{\mathbf{X}}_S^{(r)}, \mathbf{X}_{C}\}), Y\right) - \ell(\hat{f}(\mathbf{X}), Y).
\end{align}
Assuming that the distribution of $\tilde{\mathbf{X}}_S$ learned by the ARF follows the actual distribution of $\mathbf{X}_S |\mathbf{X}_{C}$, the samples \(\Delta_1, \ldots, \Delta_n\) are also i.i.d. Additionally, drawing $R$ imputations does not affect the i.i.d. condition of the averaged loss values of an instance. Thus, with a larger number of samples $N$, our cARFi estimator $\widehat{\text{cARFi}}_S = \frac{1}{N} \sum_{i = 1}^{N} \Delta_i$ converges in probability to a Gaussian distribution by the central limit theorem. Consequently, this allows for statistical significance tests as described in \citet{watson2021} (e.g., paired t-test and Fisher exact test).
However, we must note a limitation on the consistency of this convergence: the convergence is theoretically guaranteed only for models $\hat{f}$ with a finite VC dimension \citep{vapnik_1971}. Nonetheless, our simulations with support vector machines, which have an infinite VC dimension, showed consistent results (see Sec.~\ref{sec::proof-of-concept}). Additionally, uncertainties and biases in the tests may arise because the quality of the conditional distributions learned by the ARF and potential uncertainties in the machine learning model $\hat{f}$ are disregarded. 

\section{Evaluation} \label{sec::evaluation}
We evaluate the performance of cARFi on both simulated and real data. First, we demonstrate that cARFi allows for valid inference procedures and achieves high power in testing for nonzero feature importance. Next, we compare the performance of cARFi to that of CPI, its closest competitor in testing for conditional feature importance, and evaluate cARFi-based relative feature importance, both by drawing on simulation studies from previous literature. Finally, we illustrate cARFi's feature attributions to those of competing methods for a real data example.

\subsection{Proof of Concept} \label{sec::proof-of-concept}
To validate that the performance of cARFi is as expected from the theoretical considerations outlined in Sec.~\ref{sec::methods}, we draw on a simulation setup established in prior literature. Using the setup of \citet{watson2021}, we demonstrate that cARFi enables powerful and statistically valid testing for nonzero conditional feature importance. 

In detail, for $M = 10,000$ replicates, we generate $N=1,000$ instances of features $\mathbf{X} = X_1, \dots, X_{10}$ from a multivariate Gaussian distribution $\mathcal{N}(0, \Sigma)$, where $\Sigma_{ij} = 0.5^{|i-j|}$. Using effect sizes $\beta = (0.0, 0.1, \dots, 0.9)$ and additive noise $\epsilon \sim \mathcal{N}(0, 1)$, we construct target variable $Y$ according to two different settings:
%\begin{enumerate}
(1) linear setting: $\mathbf{Y} = \beta \mathbf{X} + \epsilon$;
(2) non-linear setting: $ \mathbf{Y} = \beta \mathbf{X}^{'} + \epsilon$, where $x^{'}_{ij} = 1$ if $\mathbf{\Phi}^{-1}(0.25) \leq x_{ij} \leq \mathbf{\Phi}^{-1}(0.75)$, else $x^{'}_{ij} = -1$. 
%\end{enumerate}

We fit several prediction models $\hat{f}$ to this data, including a (feedforward) neural network, support vector machine, random forest, and linear model. Subsequently, we use the mean squared error to assess $\ell$ and thus obtain test statistics.

\begin{figure}[ht]
    \centering
    \includegraphics[width=0.8\columnwidth]{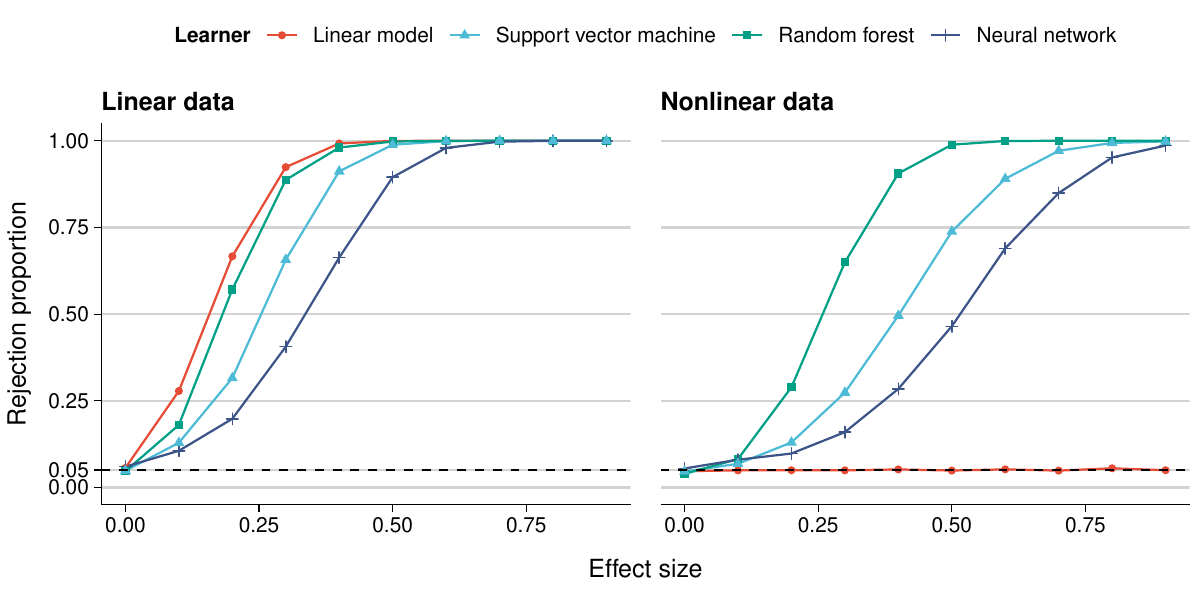}
    \caption{Rejection rates of one-sided paired $t$-tests at $\alpha ~= 0.05$ to detect relevant features at different effect sizes, i.e., type I error rates at effect size 0 and power at effect size $>$0.}
    \label{fig:poc_power}
\end{figure}

From Fig.~\ref{fig:poc_power}, we can see that cARFi acts as expected. At effect size 0, the rejection rate is at 5\%, effectively controlling type I error at the nominal level of 5\%. At positive effect sizes, power increases with effect size and all learners reach 100\% power, with the exception of the linear model on nonlinear data. 
Fig.~\ref{fig:poc_power} shows results for a minimum leaf size of 20. In addition, Supplement~S2.1 gives results for minimum leaf sizes of 2, 5, 10, 50 and 100, and further details feature importance values for the different effect sizes and empirical distributions of the test statistics. These results show that very deep trees (small leaf size) lead to slightly conservative results, i.e., type I errors below the nominal level and lower power, while shallow trees (large leaf size) show slightly inflated type I errors and higher power.

\subsection{Simulation Study} \label{sec::simulation-study}

Next, we compare the performance of cARFi to related methods in terms of time and statistical power using a simulation setting derived in previous literature.\footnote{We evaluate the performance of competing methods on simulation setups used in previous literature to promote a direct and fair comparison that is not tailored to advantageously present cARFi.} In addition, we highlight differences in feature importance measures for various conditioning sets in a simulated and real-world setting.

\subsubsection{Mixed Data} \label{sec::mixed-data}

\begin{figure}[h]
    \centering
    \includegraphics[width = 0.8\columnwidth]{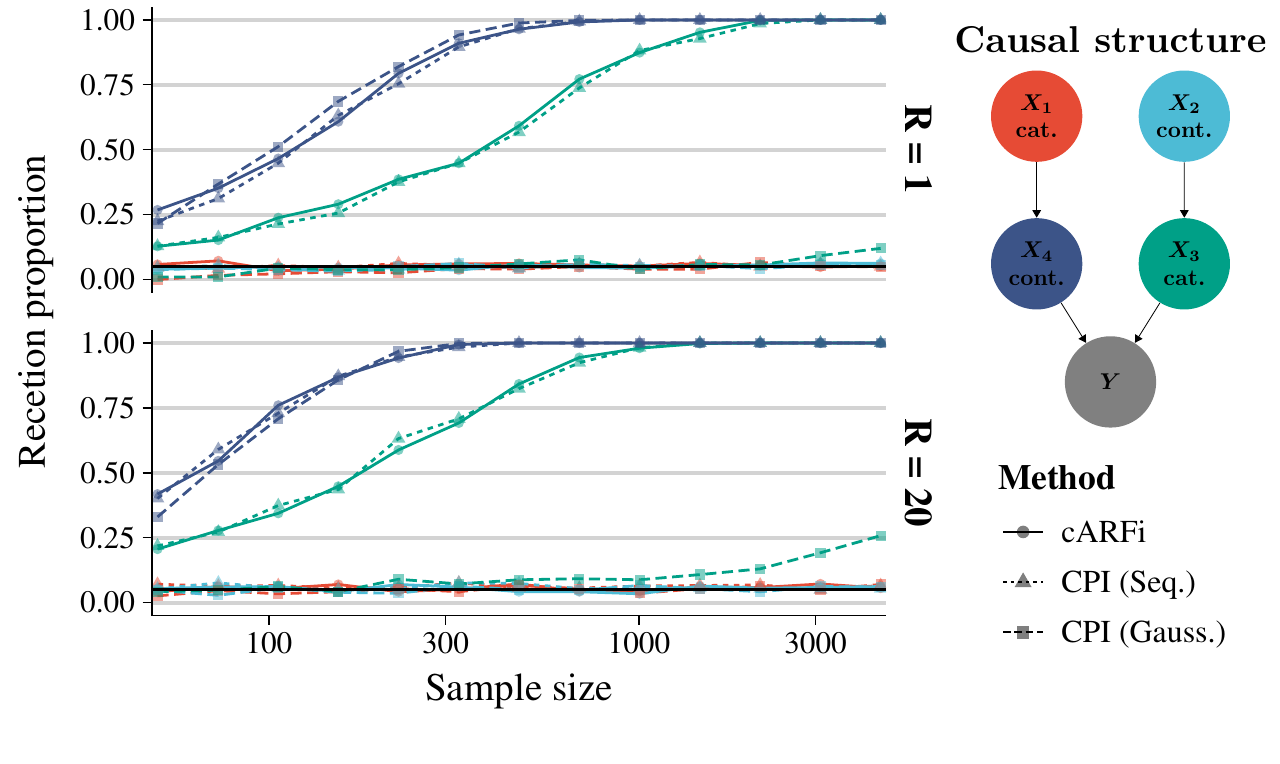}    
    \caption{Rejection rates of one-sided paired $t$-tests at $\alpha ~= 0.05$ to detect relevant features, i.e. power and type I error rates, across $500$ simulation runs. $X_1, X_3$ are 10-level categoricals, $X_2, X_4$ are Gaussian. Effect size $\beta ~= 0.5$ {  and random forest prediction model}.}
    \label{fig:DAG_Blesch}
\end{figure}

We evaluate the directed acyclic graph (DAG) introduced in scenario (III) in \citep{blesch2023} in the mixed data setting: $X_2, X_4$ are Gaussian, $X_1$ and $X_3$ are categorical with $10$ levels. Considering this DAG, the outcome $Y$ is conditionally independent of $X_1$ ($X_2$) given $X_4$ ($X_3$), or, more formally: 
$X_1 \indep Y \mid \{X_2, X_3, X_4\}$ and $X_2 \indep Y \mid  \{X_1, X_3, X_4\}$, whereas $X_3\,{\dep}\,Y \{X_1,X_2,X_4\}$  and $X_4\,{\dep}\,Y \mid \{X_1, X_2, X_3\} $. Therefore, a conditional feature importance measure should only attribute nonzero importance to variables $X_3, X_4$, but not to $X_1, X_2$. The simulation analyzes the rejection rate of the null hypothesis based on the t-test with varying sample sizes for CPI with naive dummy-encoded Gaussian knockoffs, CPI with sequential knockoffs, and cARFi. Additionally, we generate both a single knockoff or ARF-sample ($R = 1$) and multiple ones ($R = 20$), which at this point hasn't been investigated before for the CPI framework. The results using a minimum node size of 20 are shown in Fig.~\ref{fig:DAG_Blesch}.

We observe that cARFi behaves very similarly to CPI with sequential knockoffs in the sense that the type I error is mostly controlled for the conditionally unimportant features ($X_1$ and $X_2$) and shows good power for the important features ($X_3$ and $X_4$). Consistent with the findings of \citet{blesch2023}, it is notable that the naive dummy-encoded Gaussian knockoffs only correctly identify the categorical variable $X_3$ as significant at very high sample sizes. Considering the comparison with the replicates $R$ shown in the lower plot of Fig.~\ref{fig:DAG_Blesch} reveals gains in robustness. Using $R = 20$ instead of a single knockoff or ARF-sample, both power and stability of the type I error improve, especially at lower sample sizes. At this point, we want to highlight the advantage of cARFi compared to sequential knockoffs, as generating ARF samples is slightly faster with $R = 1$ and much faster with $R = 20$ than creating the corresponding knockoffs (see Supplement~S2.3).

\subsubsection{Impact of the Conditioning Set} \label{sec::additive-gaussian}

\begin{figure*}[t]
    \centering
    \includegraphics[width = 0.95\textwidth]{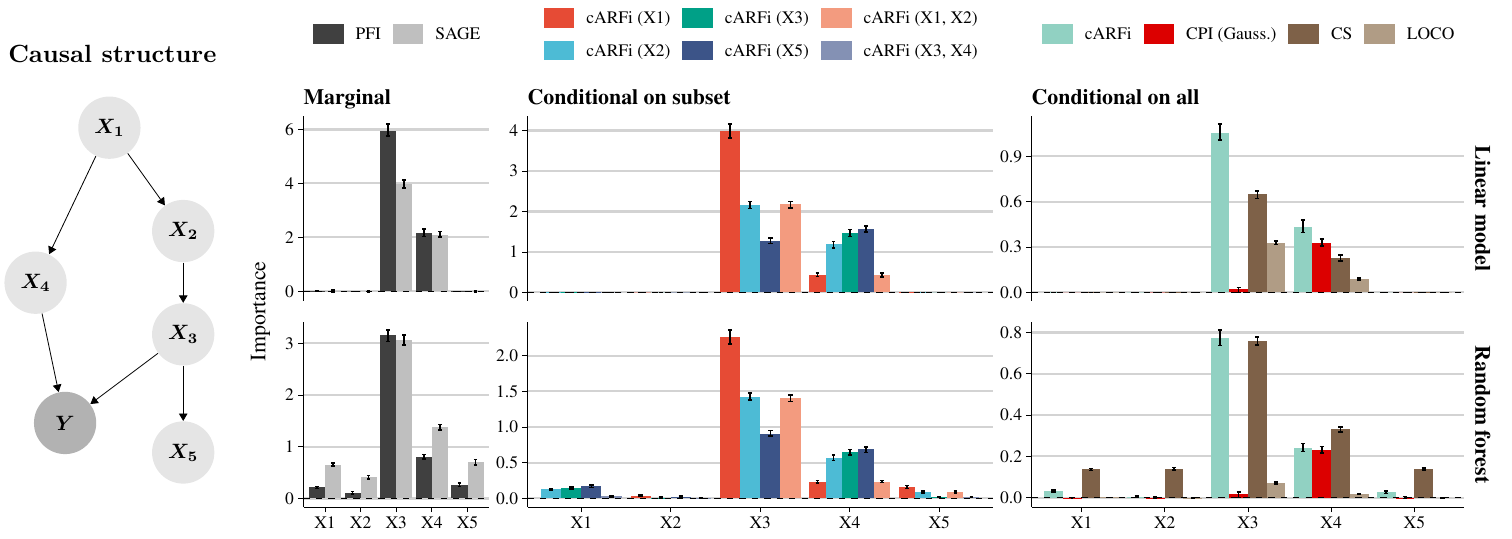}
    \caption{Marginal vs. conditional feature importances for a linear model (LM, upper row) and a random forest (RF, lower row). While the LM coefficients $\beta_1, \beta_2, \beta_5$ are close to zero, the RF assigns marginal importance to $X_1, X_2, X_5$ due to their strong correlation with $X_3, X_4$. cARFi resolves these indirect influences by conditioning on the respective feature subsets.}
    \label{fig:Cond_set}
\end{figure*}

Here, we present results for a modified version of the DAG investigated in section (VI) in \cite{koenig2021} as 
displayed in the leftmost panel of Fig.~\ref{fig:Cond_set} for two fundamentally different models.
For a random forest, the marginal measures PFI and SAGE "leak" importance from  $X_3$ and $X_4$ to the correlated features $X_1, X_2$ and $X_5$ while the linear model inherently conditions on all other features.
When we explicitly condition on all other features, the rightmost panel reveals a few interesting insights: cARFi's estimates are close to the "truth" (in a relative sense), whereas CPI assigns zero attribution to $X_3$ and CS seems unable to properly adjust its marginal measures.
Selective conditioning on $X_{1}, X_{2}$, respectively, lowers the score for $X_3$: $\widehat{\text{PFI}}_{3} > \widehat{\text{cARFi}}_3(X_1) > \widehat{\text{cARFi}}_3(X_2) \approx \widehat{\text{cARFi}}_3(X_1, X_2)$.
Similarly, conditioning on $X_5$ ($X_1$) also lowers the contribution of $X_3$ ($X_4$):  $\widehat{\text{PFI}}_3 > \widehat{\text{cARFi}}_3(X_5)$ and $\widehat{\text{PFI}}_4 > \widehat{\text{cARFi}}_4(X_1)$. For completeness, we include the various conditional independence relations as well as consistency checks and other details in Supplement~S1.

\subsection{Bike-Sharing Dataset} \label{sec::real-data}

Finally, we evaluate the behavior of cARFi under different conditioning sets in a real-world setting using the widely used bike-sharing dataset \citep{FanaeeT2013}. The dataset contains hourly records of bike rentals and includes seasonal and meteorological information such as season, weekday, humidity, and temperature. Due to the number of possible combinations of the variables, we limit our analysis to hour and temperature, which are marginally two of the most important features for the trained random forest. Even though we don't know the exact feature importance or the underlying DAG for the causal associations, we can infer some relationships based on natural and logical laws. For example, we know that the season affects the temperature, as it is colder in winter than in summer.

\begin{figure*}[t]
    \centering
    \includegraphics[width=0.9\textwidth]{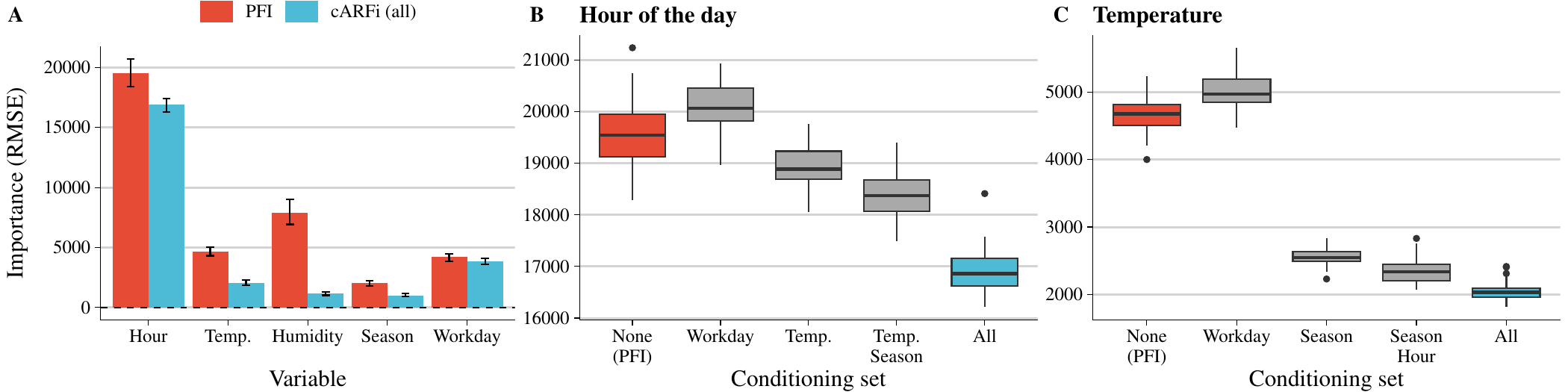}
    \caption{Feature importance values for the variables from the real-world bike rental example. Panel A: PFI and cARFi (conditioned on all other ones) values of all included variables in the random forest model. Panels B and C: cARFi values for \textit{Hour} and \textit{Temperature} for selected conditioning sets, respectively. The RMSE is used as the loss function and 50 repetitions.}
    \label{fig:bike_sharing}
\end{figure*}

We train a random forest on two-thirds of the 8\,645 instances and use the remaining as a holdout for the XAI method. We repeat this setting 50 times and apply PFI (marginal measure, i.e., no conditions), and cARFi with different conditioning sets. For cARFi, we use a minimum node size of 20, $R = 5$, and the root mean squared error (RMSE) as a loss function. The results are presented in Fig.~\ref{fig:bike_sharing}.

The left panel of Fig.~\ref{fig:bike_sharing} shows the importance values from PFI as a marginal feature importance measure and cARFi conditioned on all other features. We observe that cARFi often results in a decrease in importance as some effects can be explained through the other features, i.e., they are not conditionally independent. For example, \textit{Hour}, \textit{Temperature}, and \textit{Humidity} show a considerable drop in importance when comparing PFI to the conditional variants. In contrast, the binary variable \textit{Workday} shows only a minor change, which aligns with the intuitive understanding that whether a day is a workday or a holiday is not influenced by seasonal, temporal, or weather conditions. The other two panels illustrate the varying variable importance under different conditions.
%\footnote{We are still explaining the fitted random forest models, assuming that these models have captured the data's relationships well enough to align with natural laws or intuitive reasoning.}. 
For instance, we observe that the importance of both \textit{Hour} and \textit{Temperature} increases when conditioned only on \textit{Workday}. This is likely due to the interaction between these variables and \textit{Workday} \citep{pmlr-v206-hiabu23a}. For example, on holidays, bikes are rented at different times compared to weekdays when people commute to work in the morning. Additionally, we observe that the effect of \textit{Hour} on the number of rented bikes decreases when conditioned on \textit{Temperature}, and even more when also conditioned on \textit{Season}. This occurs because the time of day influences temperature, so the effect of \textit{Hour} through \textit{Temperature} on the bike rentals is likely partially absorbed by this condition. The reduction becomes even more apparent when \textit{Season} is added, as it captures broader trends like typical temperature ranges and daylight hours, which are closely tied to both time of day and bike usage, further decreasing \textit{Hour}'s importance for the prediction. We observe a similar effect with \textit{Temperature} when we condition on \textit{Season}: Since these two are strongly correlated, conditioning on \textit{Season} alone drastically reduces \textit{Temperature}'s importance in predicting bike rentals. This effect is further amplified when we also condition on \textit{Hour}, as it correlates with daily temperature changes, capturing much of the variability that \textit{Temperature} would otherwise explain.

\section{Conclusion} \label{sec::conclusion}
This paper presents cARFi, a method that leverages a straightforward generative model for the measurement of conditional and relative feature importance. cARFi offers robust and flexible calculation of feature importance in a model-agnostic way, and directly handles mixed data thanks to the random forest based generative procedure.  The procedure can adapt to feature subset conditioning without having to rerun the procedure and further introduces an opportunity to smoothly shift between notions of marginal, in-between and conditional feature importance measurement. In both simulated and empirical examples, the method demonstrates competitive results, and the ease of application is particularly appealing for empirical usage. An implementation of cARFi in \texttt{R} and code for reproducing the results of this paper is available at \url{https://github.com/bips-hb/cARFi_paper}.

\section{Discussion} \label{sec::discussion}
%While the paper analyses various aspects with respect to the proposed cARFi method, the paper is limited in several ways.
Even though the robustness of cARFi is analyzed regarding the number of sampled instances, see Fig.~\ref{fig:DAG_Blesch}, it may be further studied. For example, by systematically varying the sample size, dimensionality, underlying distribution of the data and choice of hyperparameters, such as the minimum node size within the ARF subroutine. Delimiting the scope of the paper, we leave such analyses for future research, yet want to highlight the necessity of providing users with comprehensive studies on such considerations. However, the relevance of parameters to focus on will highly depend on the field of application and hence, should be investigated for the specific use case at hand.

That said, studies showcasing the empirical use of cARFi through applied use cases are highly desirable. This paper focuses on the methodological proposition, hence introduces, analyzes and discusses cARFi from an abstract, general standpoint. However, cARFi is designed to facilitate conditional and relative feature importance measurement in real-world applications. Therefore, we encourage applied researchers to challenge the usefulness of cARFi in practice. 

In future work, advancements in the rapidly changing fields of generative modeling, XAI and closely related fields could be taken into account and innovative methods developed accordingly. This includes propositions that bridge other XAI methods with generative modeling, e.g., exploiting the fast subset conditioning of ARF for methods such as SHAP \citep{lundberg2017} and SAGE \citep{covert2020} that heavily rely on such operations. In principle, any well-fitted generative model that can synthesize values in accordance to the requested conditional distributions can work in such subroutines. As another example, cARFi's conditional independence testing procedure may facilitate applying algorithms in causal structure learning. Hence, cARFi can serve as a starting point for future research to propose new algorithms in various academic fields. 

\section*{Acknowledgements}
This work was supported by the German Research Foundation (DFG), grant numbers 437611051 and 459360854 and by the U Bremen Research Alliance/AI Center for Health Care, financially supported by the Federal State of Bremen. We thank Sophie Langbein for valuable discussions. Experiments were run on the  Beartooth Computing Environment \citep{beartooth}. 

\bibliography{references.bib}

\clearpage

\renewcommand{\appendixpagename}{Supplementary Material}
\begin{appendices}

\renewcommand{\thefigure}{S\arabic{figure}}
\renewcommand{\thesection}{S\arabic{section}}

\section{Impact of Conditioning Set}

Here, we supplement the results from Sec.~\ref{sec::simulation-study}  and Fig.~\ref{fig:Cond_set} with details on parameter choices and selected consistency checks.

\noindent
%\subsection{Parameter Settings}
\textbf{Parameter Settings}: 
Feature importance values shown in Fig.~\ref{fig:Cond_set} are averaged over 50 runs.
For each run, a sample of size $n=3000$ is drawn from the distribution induced by the structural causal model depicted in Fig.~\ref{fig:Cond_set}. All relationships are additive linear with coefficients 1 and Gaussian noise terms $\left(\sigma_1=\sigma_2=\sigma_3=1, \sigma_4=0.3, \sigma_5=0.7\right.$ and $\left.\sigma_y=0.5\right)$. For the cARFi hyperparameter, we use a minimum node size of 20 and a single sample, i.e., $R = 1$.

\noindent
%\subsection{Conditional Independence Relations}
\textbf{Conditional Independence Relations}: 
Considering this DAG, the outcome $Y$ is conditionally independent of $X_1$, $X_2$ and $X_5$ given $X_4$ and $X_3$, or, more formally: 
$X_1 \perp\!\!\!\perp Y \mid \{X_3, X_4\}$ and $X_2 \perp\!\!\!\perp Y \mid  \{X_3, X_4\}$, whereas $X_3 \not\!\perp\!\!\!\perp Y \{X_1,X_2,X_4,X_5\}$  and $X_4  \not\!\perp\!\!\!\perp Y \mid \{X_1, X_2, X_3,X_5\} $. Therefore, a conditional feature importance measure should only attribute nonzero importance to variables $X_3, X_4$, but not to $X_1, X_2, X_5$.

\noindent
%\subsection{Consistency Checks}
\textbf{Consistency Checks}: 
\begin{itemize}
    \item \cite{gregorutti2015grouped} derived an explicit analytic result for the expected value of PFI for a linear model of the form $f(\mathbf{X})=\hat{\beta}_0+\sum_{j=1}^p \hat{\beta}_j X_j$:
\[
\mathbb{E}_\pi [\text{PFI}_j^\pi] \approx 2 \hat{\beta}_j^2 \cdot \frac{1}{n} \sum_{i=1}^n\left(x_{i j}-\bar{x}_j\right)^2 \approx 2 \hat{\beta}_j^2 \cdot \mathrm{Var}(X_j)
\]
where $\pi$ denotes the PFI permutations.% and $\bar{x}_j$ is the average value of the $j$th feature.
The variances for $X_3, X_4$ are $3$ and $1.09$, respectively.
With $\beta_3 = \beta_4 = 1$, we would expect the theoretical PFIs to be 
$\text{PFI}_3 = 2 \cdot 3 = 6$ and 
$\text{PFI}_4 = 2 \cdot 1.09 = 2.18$. Those values are very close to the empirical PFI scores in the upper "Marginal" panel of Fig. 3.
    \item Once we condition on $X_1$, the FI score for $X_4$ should not change by further conditioning. Indeed, we have $\widehat{\text{cARFi}}_4(X_1) \approx \widehat{\text{cARFi}}_4(X_1, X_2) \approx \widehat{\text{cARFi}}_4(\text{all}) \approx 0.4$ from Fig. 3.
    \item Once we condition on $X_2$, the FI score for $X_3$ should not change by further conditioning on any variable besides $X_5$. Indeed, we have $\widehat{\text{cARFi}}_3(X_2) \approx \widehat{\text{cARFi}}_3(X_1, X_2)  \approx 2.2$ from Fig. 3.
\end{itemize}

\section{Additional Figures}

The following figures supplement the results of Sec. 4. All experimental setups are identical to the ones described in the main document.

\subsection{Proof of Concept: Effect of Minimum Node Size}

The following figures show the impact of the hyperparameter minimum node size and are structured as follows, respectively: 
\begin{itemize}
    \item Panel A: Boxplots of feature importance values for increasing effect sizes. 
    \item Panel B: Distributions of test statistics of variables with effect size 0. The distribution of the expected t-statistic under the null hypothesis is shown in red. 
    \item Panel C: Proportion of rejected hypotheses at $\alpha = 0.05$ as a function of effect size. Results at effect size 0 correspond to the type I error, at positive effect sizes to statistical power. The dashed line indicates the nominal level, $\alpha = 0.05$.
\end{itemize}

\begin{figure}[H]
    \centering
    \includegraphics[width=0.975\linewidth]{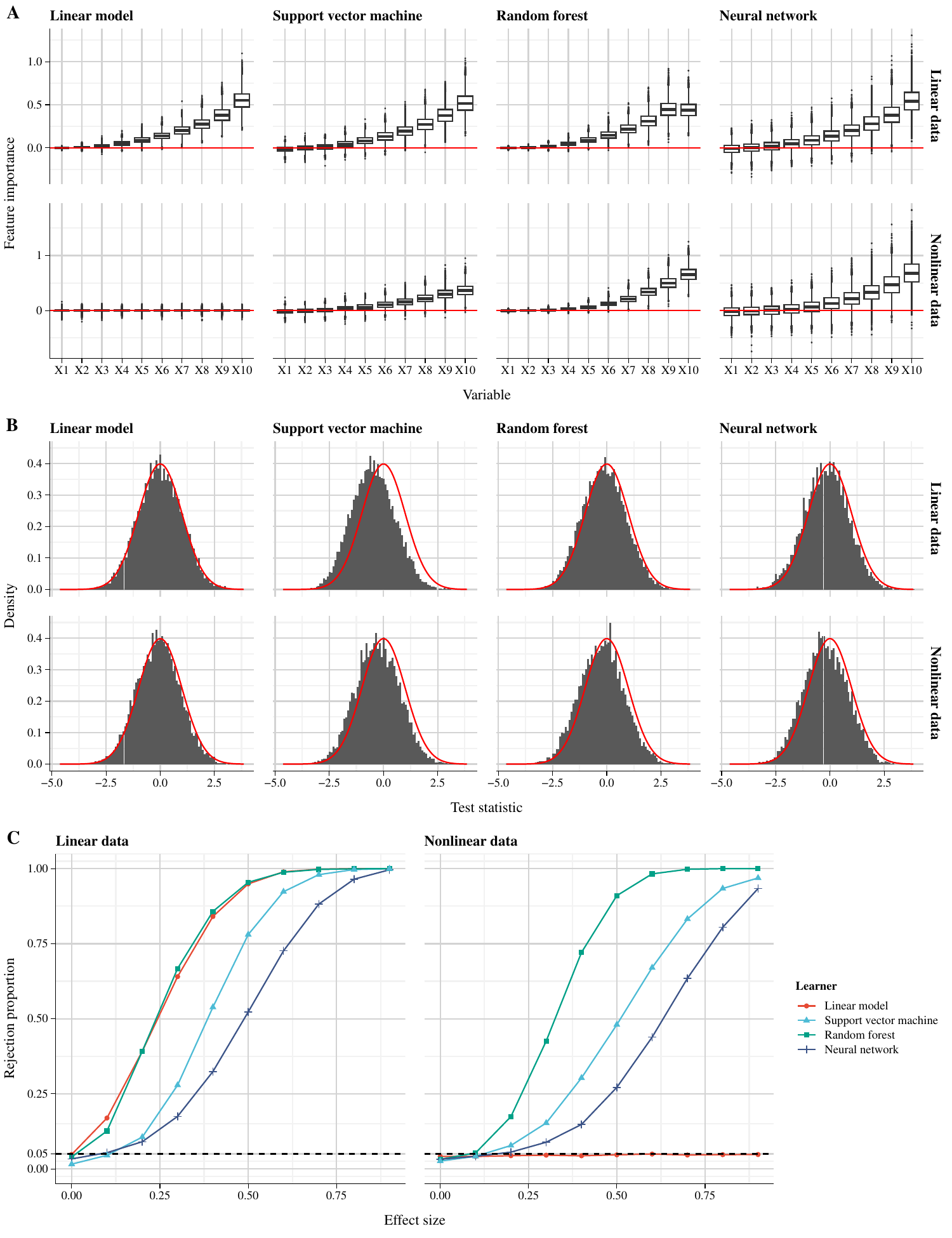}
    \caption{Proof of concept for minimum leaf size of 2.}
    \label{fig:poc2}
\end{figure}

\begin{figure}[H]
    \centering
    \includegraphics[width=\linewidth]{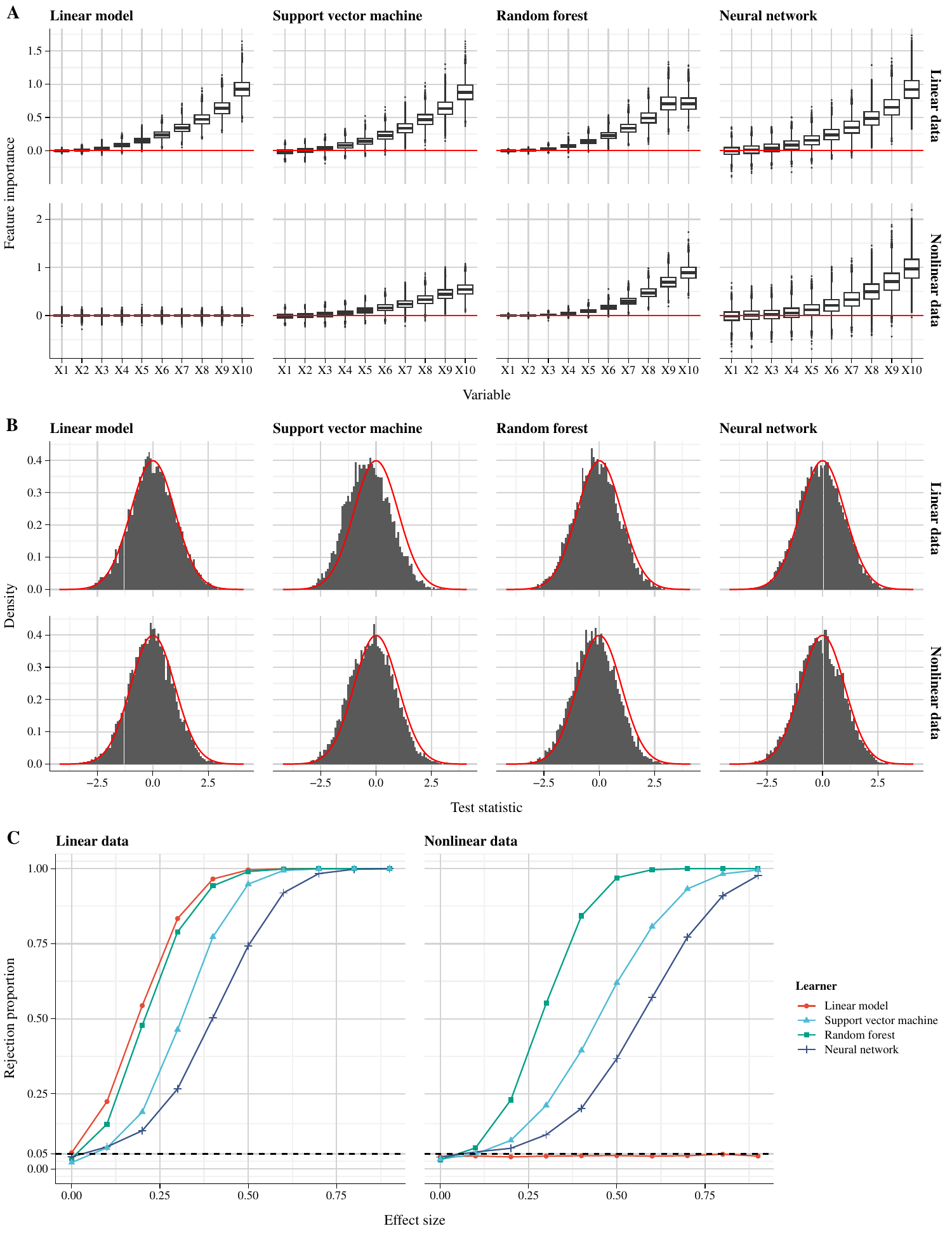}
    \caption{Proof of concept for minimum leaf size of 5.}
    \label{fig:poc5}
\end{figure}

\begin{figure}[H]
    \centering
    \includegraphics[width=\linewidth]{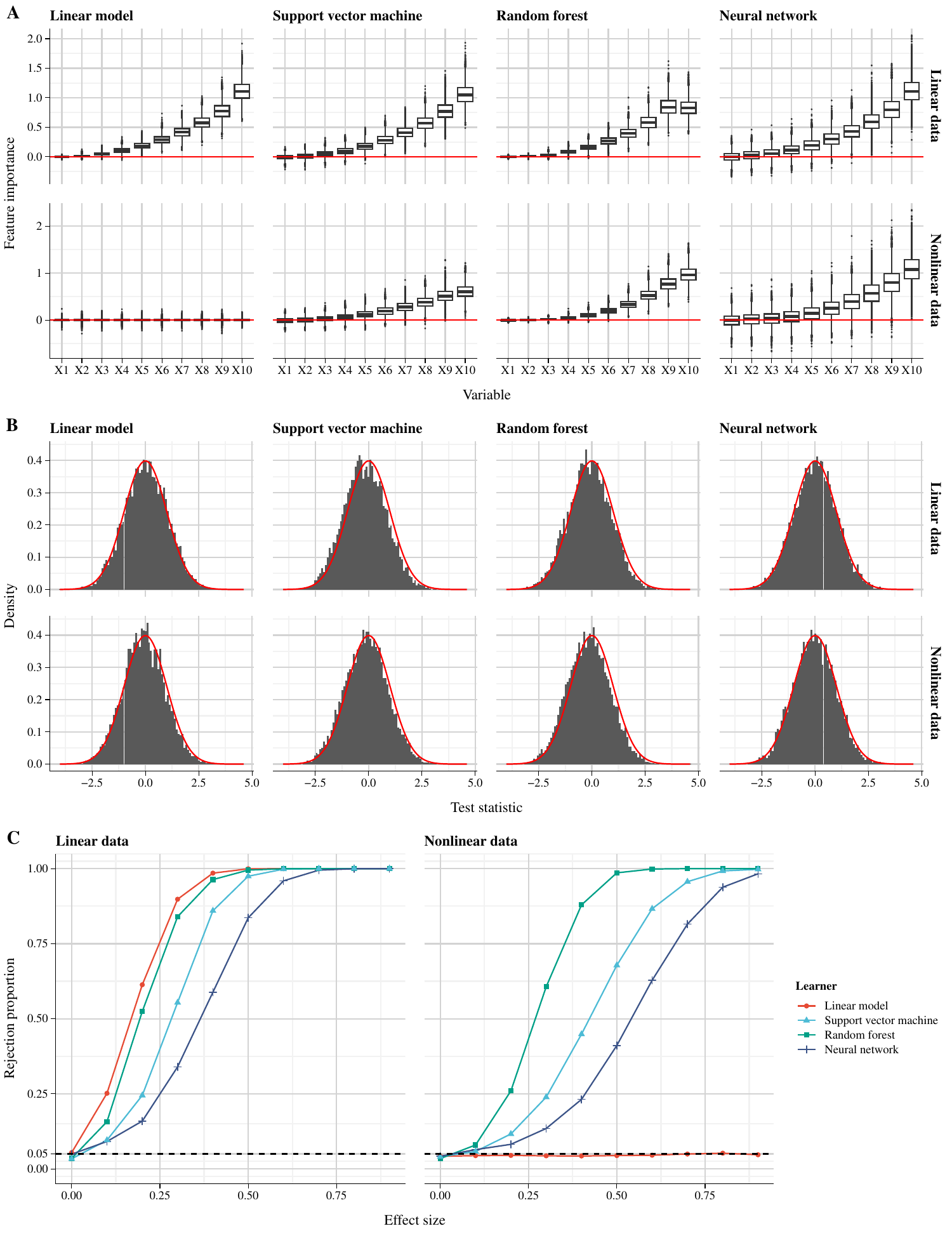}
    \caption{Proof of concept for minimum leaf size of 10.}
    \label{fig:poc10}
\end{figure}

\begin{figure}[H]
    \centering
    \includegraphics[width=\linewidth]{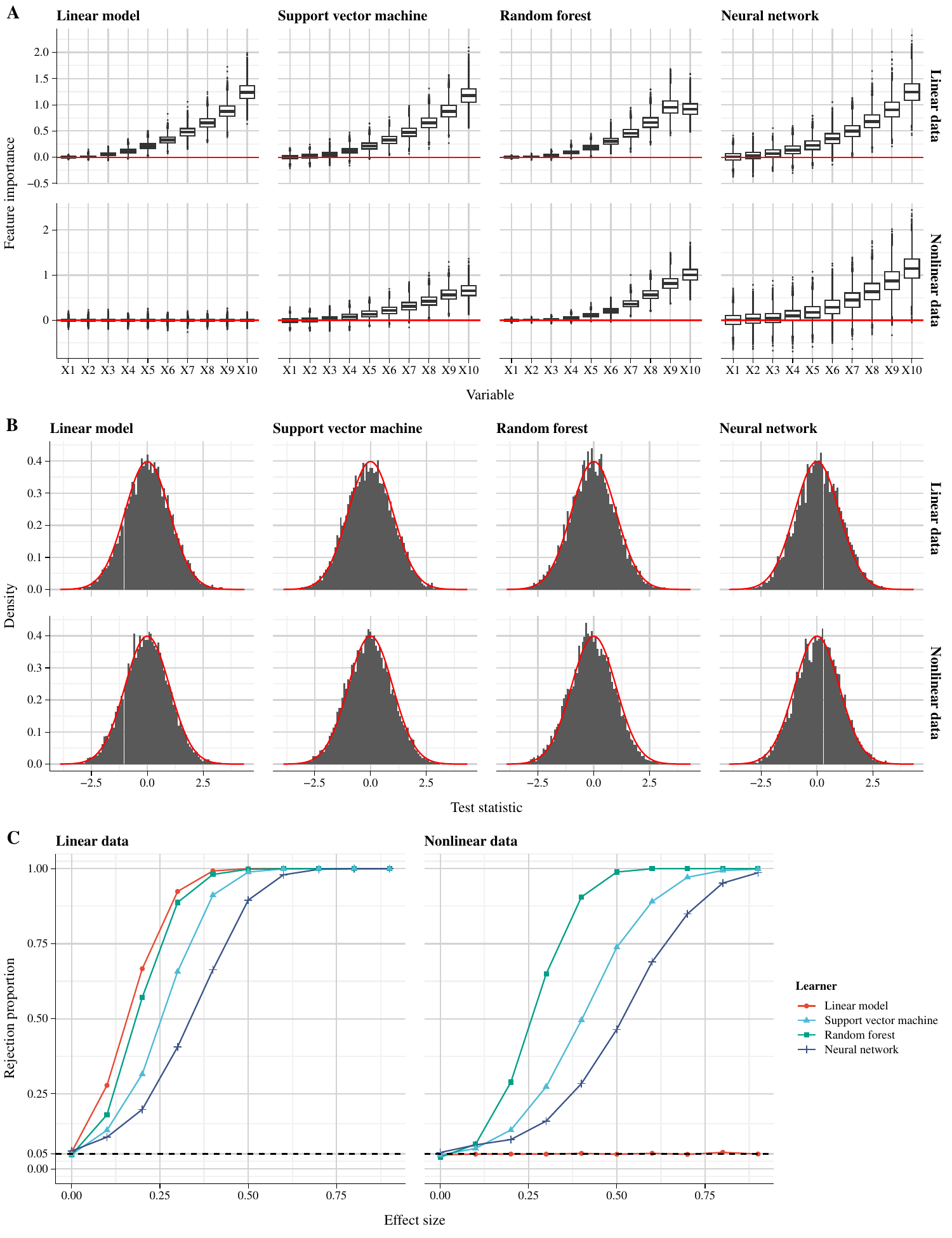}
    \caption{Proof of concept for minimum leaf size of 20.}
    \label{fig:poc20}
\end{figure}

\begin{figure}[H]
    \centering
    \includegraphics[width=\linewidth]{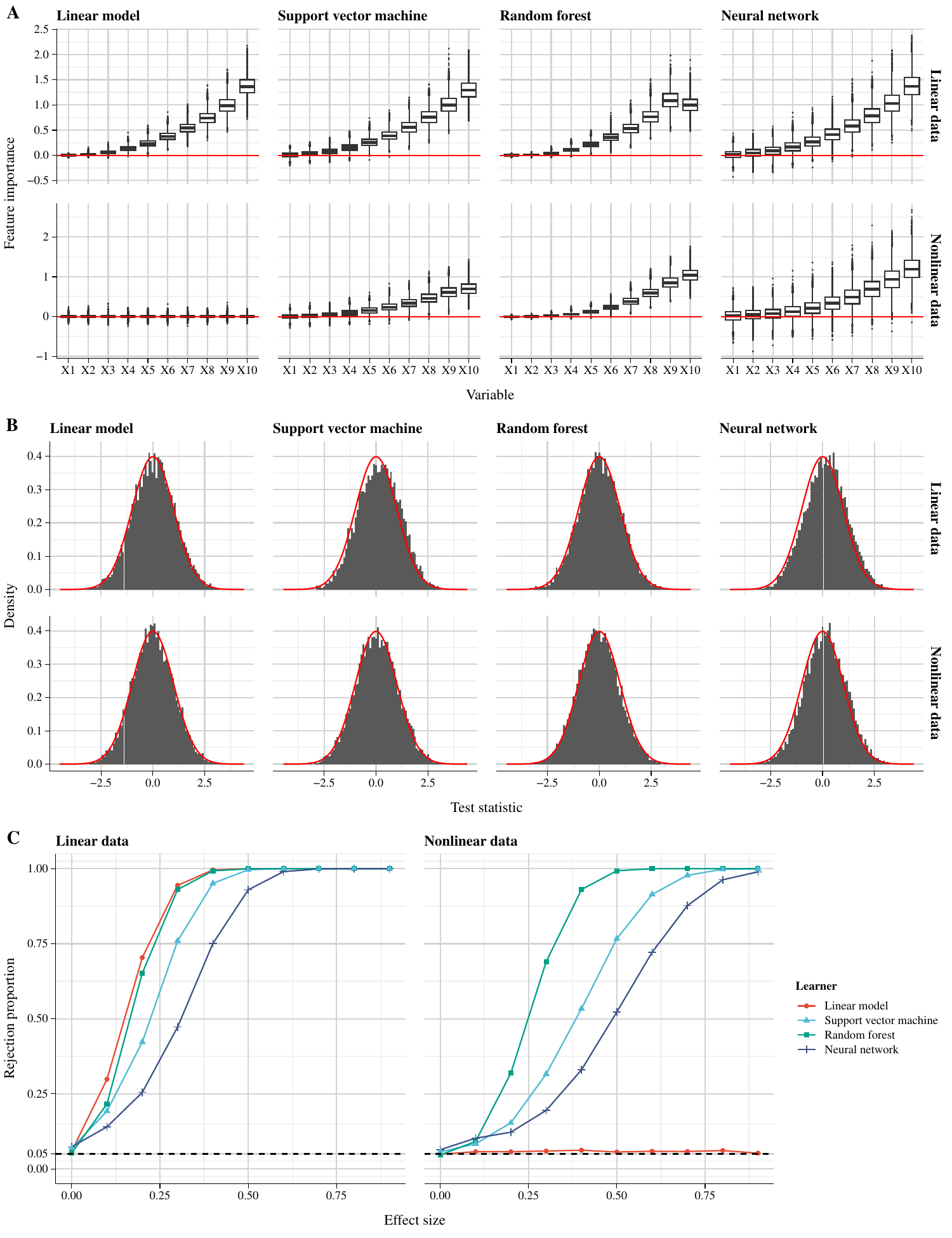}
    \caption{Proof of concept for minimum leaf size of 50.}
    \label{fig:poc50}
\end{figure}

\begin{figure}[H]
    \centering
    \includegraphics[width=\linewidth]{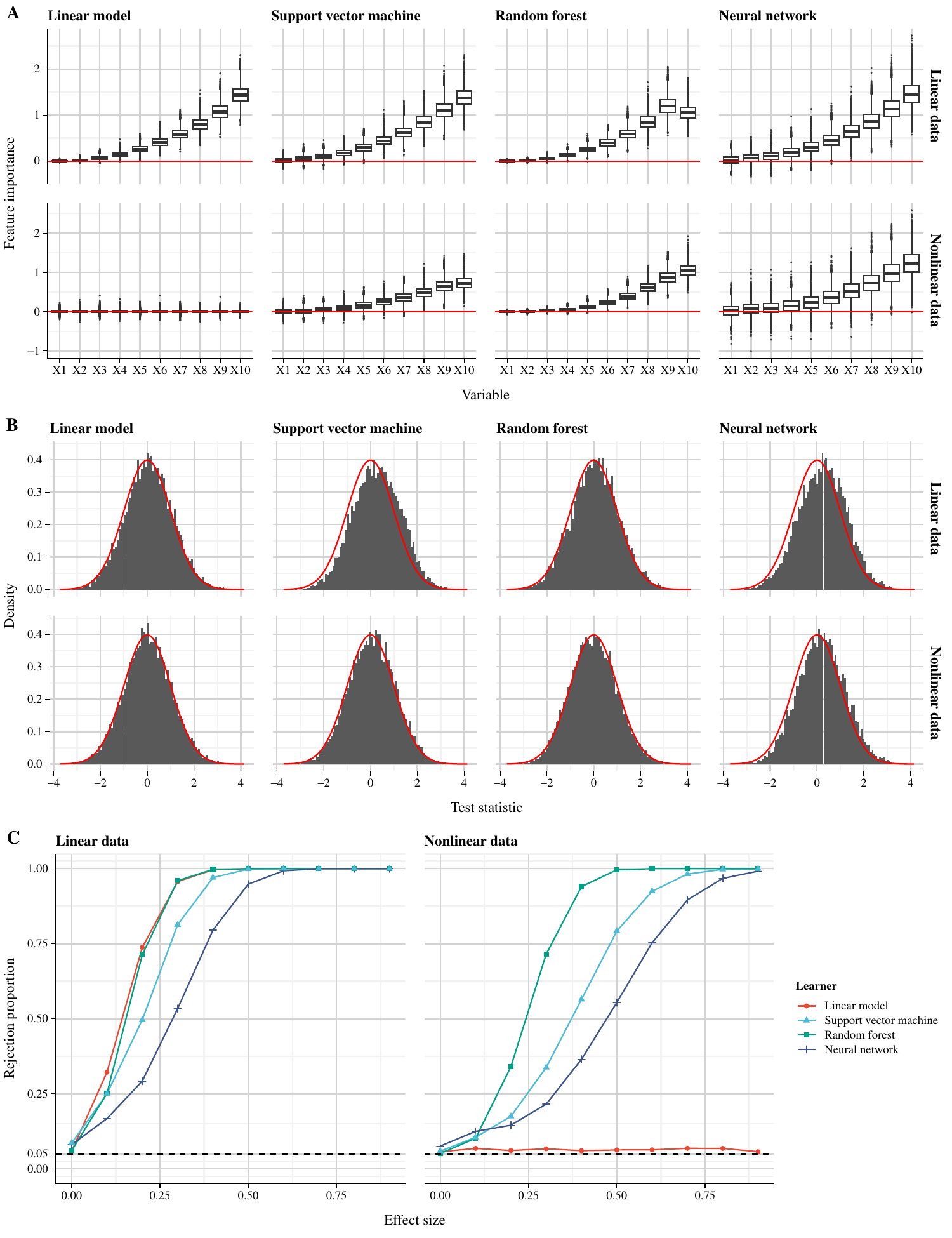}
    \caption{Proof of concept for minimum leaf size of 100.}
    \label{fig:poc100}
\end{figure}

\subsection{Mixed Data Simulation: All Continuous Case}

\begin{figure}[H]
    \centering
    \includegraphics[width=0.975\linewidth]{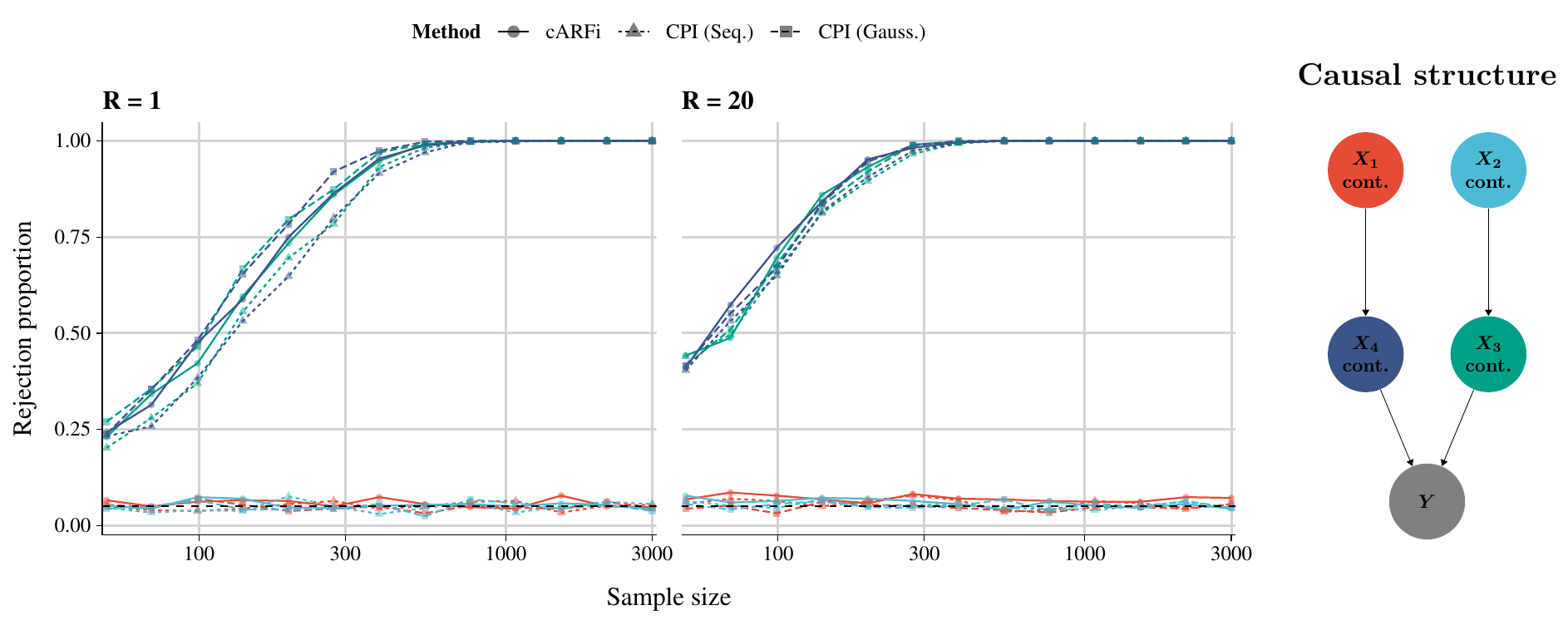}
    \caption{Rejection rates of one-sided paired $t$-tests at $\alpha ~= 0.05$ to detect relevant variables, i.e. power and type I error rates, across $500$ simulation runs. All variables $X_1, \ldots, X_4$ are Gaussian. Effect size $\beta ~= 0.5$ and random forest prediction model.}
    \label{fig:mixed_data_cont}
\end{figure}

\subsection{Mixed Data Simulation: Runtime Comparison}

\begin{figure}[H]
    \centering
    \includegraphics[width=0.975\linewidth]{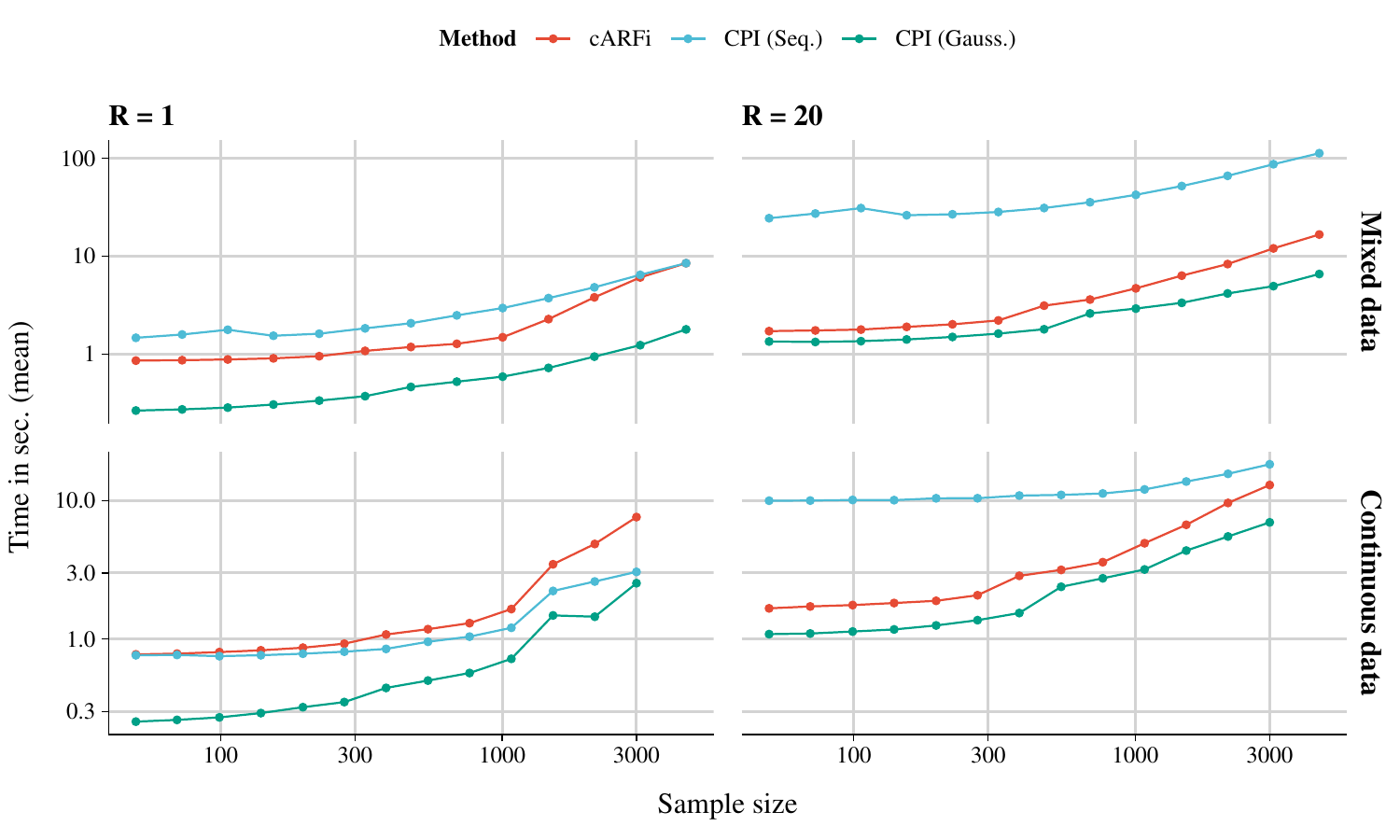}
    \caption{Computational runtime comparison in seconds of cARFi and the competing CPI method using sequential and Gaussian knockoffs on the mixed data simulation setting described in Sec.~2.2. Each method is executed in parallel on 5 threads.}
    \label{fig:mixed_data_time}
\end{figure}

\end{appendices}

\end{document}